\documentclass[11pt,a4paper]{article}
\usepackage[hyperref]{emnlp2020}
\usepackage{times}
\usepackage{soul}
\usepackage{url}
\usepackage{hyperref}
\usepackage[utf8]{inputenc}
\usepackage{caption}
\usepackage{graphicx}
\usepackage{amsmath}
\usepackage{amssymb}
\usepackage{amsthm}
\usepackage{booktabs}
\usepackage{algorithm}
\usepackage{algorithmicx}
\usepackage{algpseudocode} 
\usepackage{multirow}
\usepackage{tabularx}
\usepackage{booktabs}
\usepackage{url}
\usepackage[normalem]{ulem}
\usepackage{microtype}
\usepackage{latexsym}

\usepackage{array}
\usepackage{amsmath, bm}
\usepackage{amssymb}
\usepackage{dsfont}
\usepackage{varwidth}
\usepackage{enumitem}
\usepackage{booktabs}
\usepackage{stmaryrd}
\usepackage{microtype}

\aclfinalcopy 

\title{Program Enhanced Fact Verification with Verbalization and \\Graph Attention Network}

\author{
Xiaoyu Yang$^\dagger$\footnotemark[1],
~~Feng Nie$^\S$\footnotemark[1],
~~Yufei Feng$^\dagger$,  
~~Quan Liu$^\ddagger$, 
~~\textbf{Zhigang Chen}$^\ddagger$, 
~~\textbf{Xiaodan Zhu}$^\dagger$ \\ 
  $^\dagger$ ECE \& Ingenuity Labs Research Institute, Queen's University \\ 
  $^\S$ 
    Sun Yat-sen University
  \\ 
  $^\ddagger$ State Key Laboratory of Cognitive Intelligence, iFLYTEK Research \\
  \texttt{\{xiaoyu.yang, xiaodan.zhu, yufei.feng\}@queensu.ca} \quad \\ \texttt{fengniesysu@gmail.com} 
  \texttt{\{quanliu, zgchen\}@iflytek.com}
  }

\date{}
\begin{document}
\maketitle

\renewcommand{\thefootnote}{\fnsymbol{footnote}} 
\footnotetext[1]{Equal contribution to this work. The work was done during the second author's visiting to Queen's University.}

\begin{abstract}
Performing fact verification based on structured data is important for many real-life applications and is a challenging research problem, particularly when it involves both symbolic operations and informal inference based on language understanding. In this paper, we present a \textbf{Prog}ram-enhanced \textbf{V}erbalization and \textbf{G}raph \textbf{AT}tention Network (\textbf{ProgVGAT}) to integrate \textit{programs} and \textit{execution} into textual inference models.
Specifically, a \textit{verbalization with program execution} model is proposed to accumulate evidences that are embedded in operations over the tables. Built on that, we construct the \textit{graph attention verification networks}, which are designed to fuse different sources of evidences from verbalized program execution, program structures, and the original statements and tables, to make the final verification decision. To support the above framework, we propose a \textit{program selection} module optimized with a new training strategy based on margin loss, to produce more accurate programs, which is shown to be effective in enhancing the final verification results. Experimental results show that the proposed framework achieves the new state-of-the-art performance, a 74.4\% accuracy, on the benchmark dataset TABFACT. 
Our code is available at https://github.com/arielsho/Program-Enhanced-Table-Fact-Checking.

\end{abstract}

\section{Introduction}
\begin{figure}[t]
  \centering
  \includegraphics
[width=\linewidth,trim={0cm 9cm 17.5cm 0cm},clip]
{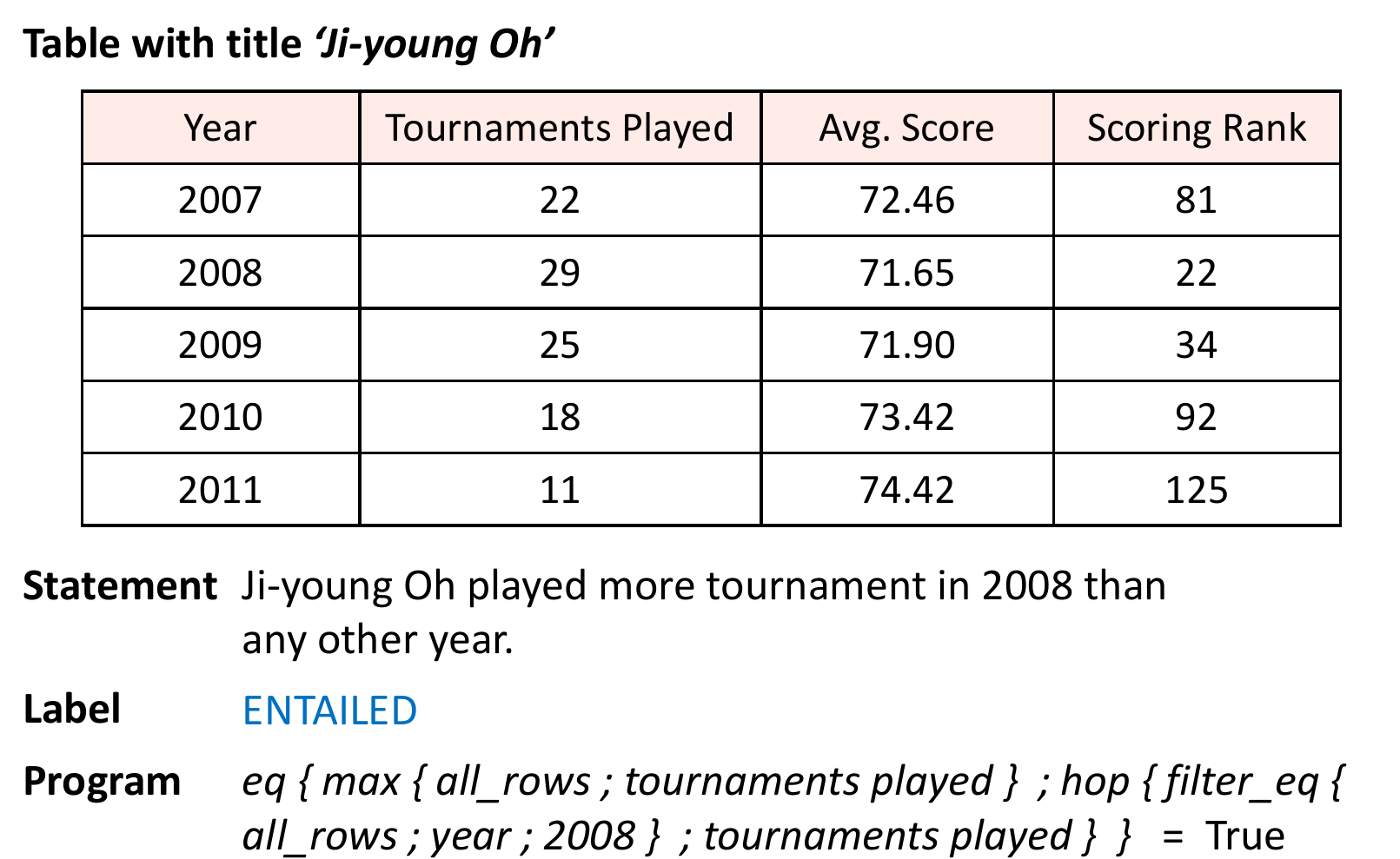}
  \caption{An example of fact verification over tables.}
\label{fig:example-1}
\end{figure}

With the overwhelming information available on the Internet, fact verification has become crucial for many applications such as detecting fake news, rumors, and political deception~\cite{rashkin2017truth, thorne2018fever, goodrich2019assessing, vaibhav2019sentence, kryscinski2019evaluating}, among others. 
Existing research has mainly focused on collecting and processing evidences from unstructured text data~\cite{liu2019finegrained,nie2019combining, hanselowski2018ukp, yoneda2018ucl}, which is only one type of data where important facts exist. Structured and semi-structured data, e.g., tables in relational databases or in the HTML format is also ubiquitous.
Performing fact validation based on structured data is important yet challenging and further study is highly desirable. Fig.~\ref{fig:example-1} depicts a simplified example in which systems are expected to decide whether the facts in the table support the natural language statement. 

In addition to its importance in applications, the task presents research challenges of fundamental interests---the problem inherently involves both informal inference based on language understanding~\cite{ dagan2005pascal,maccartney2009natural, maccartney2008modeling,bowman2015large, bowman2016fast}
and symbolic operations such as mathematical operations (e.g., \textit{count} and \textit{max}). 
Recently, pre-trained language models such as BERT ~\cite{devlin2018bert} have shown superior performances in natural language inference by leveraging knowledge from large text datasets and can capture complicated semantic and syntactic information among premises and hypotheses~\cite{radford2018improving, radford2019language, liu2019roberta, dong2019unified}.
However, such methods tend to fail when verification requires the joint modelling of both symbolic operations and language inference ~\cite{GuptaLR0020} such as the case depicted in Fig.~\ref{fig:example-1}. 

To effectively enable symbolic operations and integrate them into language-based inference models, we propose a framework centered around programs, i.e., logical forms that can be executed to find evidences from structured data. Our model starts with a \textit{program selection} module, for which we propose a new training strategy based on margin loss to find programs that can accurately extract supporting facts from tables. To bridge the semantic gap between structured programs and tables as well as to leverage the structures of programs, we propose a novel model based on \textit{verbalization with program execution}. The verbalization algorithm interweaves with program execution in order to accumulate evidences inherently embedded in operations, and the algorithm recursively converts executed operations in programs into natural language sentences. Built on that, 
we propose \textit{graph-based verification network} to fuse different sources of evidences from verbalized program execution, together with the original statements and tables, to support the final verification decision. 

We conduct experiments on the recently proposed large scale benchmark dataset TABFACT~\cite{chen2019tabfact}. Experimental results show that our proposed framework achieves new state-of-the-art performance, an accuracy of 74.4\%, substantially improving the previously reported best performance with the accuracy of 71.7\%. Our detailed analysis shows the effectiveness of verbalization and graph-based verification network in utilizing programs to achieve the improvement. The analysis also demonstrates that the program selection optimized with the proposed training strategy based on margin loss effectively improves the final verification results. 

\begin{figure*}[htb]
  \centering
  \includegraphics
  [width=\linewidth,trim={1cm 0.5cm 1cm 1cm},clip]
   {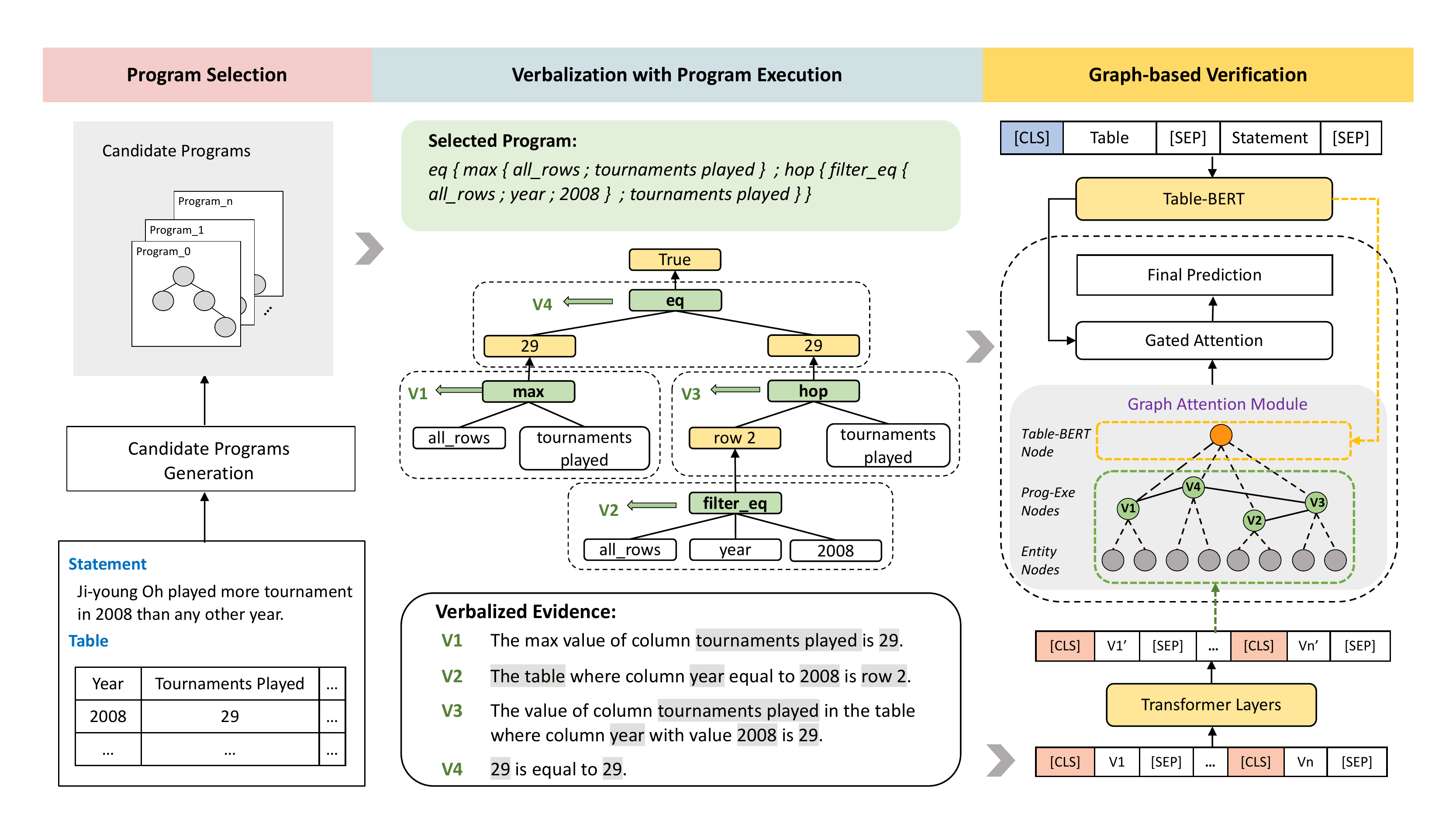}
  \caption{An overview of the proposed framework.} 
\label{fig:framework}
\end{figure*}

\section{Related Work}
\label{relatedwork}
\textbf{Fact Verification.} Existing work on fact verification is mainly based on collecting and using evidences from unstructured text data ~\cite{liu2019finegrained,nie2019combining, hanselowski2018ukp, yoneda2018ucl}. FEVER~\cite{thorne2018fever} is one of the most influential benchmark datasets built to evaluate systems in checking claims by retrieving Wikipedia articles and extracting evidence sentences. 
Recent proposed FEVER 2.0~\cite{thorne2019fever2} has a more challenging dataset to verify factoid claims and an adversarial attack task. Some previous models are developed on the official baseline~\cite{thorne2018fever} with three step pipeline~\cite{chen2017-reading} for fact verification~\cite{hanselowski2018ukp,yoneda2018ucl,yin-2018-twowingos,nie2019combining}. Others formulates fact verification as graph reasoning~\cite{zhou-2019-gear,liu2019finegrained}. 
Natural language inference (NLI) task is also a verification problem which is fully based on unstructured text data~\cite{dagan2005pascal,dagan2010recognizing,bowman2015large,parikh-2016-decomposable,chen-2017-enhanced,ghaeini-2018-dr,peters-2018-deep}. Neural models proposed for NLI have been shown to be effective~\cite{parikh-2016-decomposable,chen2016enhanced,chen2017recurrent,ghaeini-2018-dr,peters-2018-deep}, including models incorporating external knowledge~\cite{chen2017neural,yang2019enhancing}. Our work focuses on fact verification based on structured tables~\cite{chen2019tabfact}.

For verification performed on structured data, ~\citet{chen2019tabfact} propose a typical baseline (Table-BERT), which is a semantic matching model taking a linearized table $T$ and statement $S$ as input and employs BERT for verification. The other model (LPA) proposed in~\cite{chen2019tabfact} uses Transformer blocks to compute semantic similarity between a statement and program. 
A contemporaneous work~\cite{zhong2020logicalfactchecker}
proposes LogicalFactChecker aiming to leverage programs for fact verification. LogicalFactChecker utilizes inherent structures of programs to prune irrelevant information in evidence tables and modularize symbolic operations with module networks. Different from theirs, our proposed framework verbalizes the accumulated evidences from program execution to support the final verification decision with graph attention networks.



\noindent
\textbf{Semantic Parsing.} A line of work uses program synthesis or logic forms to address different natural language processing problems, such as question answering ~\cite{berant2013semantic,berant2014-semantic}, code generation ~\cite{yin-2017-syntactic}, SQL synthesis ~\cite{zhong2017seq2sql,yu2018spider} and mathematical problem solving~\cite{kushman-2014-learning,shi-2015-automatically}. Traditional semantic parsing methods greatly rely on rules and lexicons to parse texts into structured representations ~\cite{zettlemoyer2012learning,berant2013semantic,artzi2013weakly}. Recent semantic parsing methods strives to leverage the power of neural networks~\cite{NeelakantanLS15,jia2016data,liang-nsm,yu2018spider,dong2019neural}.
Our work leverages symbolic operations inherited in programs produced by neural semantic parsers to enhance fact verification over structured data.


\section{Model}
We present the proposed framework (ProgVGAT), which centers around \textit{programs} and \textit{execution} to integrate symbolic operations for fact verification. Fig.~\ref{fig:framework} depicts the overall architecture.
This section is organized as follows. We first introduce the task formulation along with program representations in Sec. \ref{sec:formulate}. Then, we describe in Sec.~\ref{sec:program selection} the program selection module that aims to obtain a semantically relevant program for verification. Built on that, we present our proposed verbalization algorithm and graph attention network to dynamically accumulate evidences embedded in symbolic operations of programs for final verification in Sec.~\ref{sec:verb} and Sec.~\ref{sec:graph-veri}.

\subsection{Task Formulation and Notations}\label{sec:formulate}
Formally, given a structured evidence table $T$ and a statement $S$, the fact verification task aims to predict whether $T$ \textit{entails} $S$ or \textit{refutes} it. The evidence table $T = \{T_{i,j} |i \leq R , j \leq C \}$ has $R$ rows and $C$ columns, and $T_{i, j}$ is the value in the $(i, j)$-th cell. $T_{i,j}$ can be of different data types, e.g., a word, number, phrase, or even natural language sentence.

\noindent
\textbf{Program representation.} Given a statement $S$, a semantic consistent program $z=\{op_i\}_{i=1}^M$ is a tree consisting of multiple executable symbolic operations $op_i$. An example of programs is shown in the center of Fig.~\ref{fig:framework}.  
An operation $op_i=(op_i.t, op_i.arg)$ contains an operator $op_i.t$ (e.g., \textit{max} in the figure) and arguments $op_i.arg$ relevant to table $T$ (e.g., \textit{all rows} and \textit{tournaments played}), and the execution of an operation yields an output/answer $ans$ (e.g., \textit{29}). Before building the model, we follow the previous work~\cite{chen2019tabfact} and perform rule-based entity linking and latent program search to obtain a set of candidate programs $\mathcal{Z}=\{z_i\}_{i=1}^{N}$. Specifically, entity linking \cite{nieaaai18} detects relevant entities (i.e., cells in evidence table $T$) in statement $S$ using a set of string matching rules. And the latent program search algorithm finds all valid combinations of pre-defined operations and detected entities by traversing and executing them recursively through the evidence table $T$.

\subsection{Program Selection}
\label{sec:program selection}
Given a statement $S$, program selection aims to obtain a high quality program $z^*$ from a set of candidate programs $\mathcal{Z}=\{z_i\}_{i=1}^{N}$.
Previous work ~\cite{chen2019tabfact} optimizes the model via a cross entropy loss:
\begin{align}
J(\theta) = -\sum_{z\in\mathcal{Z}} \mathds{1}_{{[z]_{y}}}\log p_{\theta}(z|S, T) \label{program:prob}
\end{align}
where $\mathds{1}_{{[z]_{y}}}$ is an indicator function which takes the value 1 if the execution result of a program $z$ (i.e., the output of the root; \textit{``True"} in Fig.~\ref{fig:framework}) is consistent with the ground-truth verification label $y$, otherwise 0. The former type of programs are called \textit{label-consistent} programs and the latter \textit{label-inconsistent} programs. Despite being simple, it ignores that only one of the label-consistent programs is correct and can potentially assign too much credit to spurious programs that execute to get the correct verification labels with incorrect operations during training. 
Meanwhile, the loss function considers every program in $\mathcal{Z}$ during training and there is only one most relevant program $z^*$ selected in testing phase, creating discrepancies between training and testing.


To remedy these issues, we introduce a margin loss which encourages to select a most positive program (i.e., positive program with maximum semantic similarity score) while maintaining a margin with label-inconsistent programs:
\begin{align}
    J(\theta) \!=\! \max\Big(p_{\theta}(z^{'}_{neg}|S, T) - & p_{\theta}(z^{'}_{pos}|S, T)\!+\!\gamma, 0\Big)
    \label{eqn:margin_loss} 
\end{align}
\noindent where $z^{'}_{neg}$ and $z^{'}_{pos}$ refer to the label-inconsistent program and the label-consistent program with the highest probability in their own categories, respectively. $\gamma$ is a hyperparameter controlling the margin between the positive instances and negative ones. To measure the semantic similarity between a candidate program $z$ and the corresponding statement $S$, we leverage pre-trained language model BERT \cite{devlin2018bert} instead of simply training transformer layers as proposed in~\cite{chen2019tabfact}. Specifically, given a $(S, z)$ pair, it is prefixed with the [\texttt{CLS}] token and suffixed with the [\texttt{SEP}] token to form the input of BERT. Then a 1-layer MLP with a sigmoid layer is applied on top of the BERT model to produce the final probability $p_{\theta}(z|S, T) = \sigma(W_{r}\textbf{\textit{h}})$.

Instead of selecting the top program based on Eq.~\ref{eqn:margin_loss}, we also tried the exploration strategy proposed in~\cite{guu2017language} to sample a non-top label-consistent program with a small probability. However, this does not further improve the verification performance on the development dataset. We therefore use the first ranked program in the remainder of this paper. Our proposed method relies on the program produced by this section. We further conclude the importance of the program quality in the experimental sections (i.e., Sec.\ref{sec:ablation-3}). 
\begin{table*}[t]
\small
\centering
\begin{tabular}{l|ll}
\toprule[1pt]
\multicolumn{1}{c|}{\multirow{2}{*}{\textbf{Operation}}}                                              & \multicolumn{2}{c}{\textbf{Templates}}                                                                                                                                                                                                                                                    \\ \cline{2-3} 
\multicolumn{1}{c|}{}                                                                                 & \multicolumn{1}{c|}{Operation}                                                                                                          & \multicolumn{1}{c}{Operation Results}                                                                                                           \\ \hline
count, {[}verb\_arg1{]}, ans:string/number                                                                          & \multicolumn{1}{l|}{the number of {[}verb\_arg1{]}}                                                                                     & the number of {[}verb\_arg1{]} is {[}to\_string(ans){]}                                                                                              \\ \hline
\begin{tabular}[c]{@{}l@{}}greater, {[}verb\_arg1, \\ verb\_arg2{]}, ans:true/false\end{tabular}          & \multicolumn{1}{l|}{\begin{tabular}[c]{@{}l@{}}{[}verb\_arg1{]} greater than \\ {[}verb\_arg2{]}\end{tabular}}                          & \begin{tabular}[c]{@{}l@{}}{[}arg1's template{]} {[}true:is{]}/{[}false:is not{]}\\  greater than {[}arg2's template{]}\end{tabular}                       \\ \hline
\begin{tabular}[c]{@{}l@{}}filter\_eq, {[}verb\_arg1, \\ verb\_arg2, verb\_arg3{]},  ans:rows\end{tabular} & \multicolumn{1}{l|}{\begin{tabular}[c]{@{}l@{}}{[}verb\_arg1{]} where column\\ {[}verb\_arg2{]} equal to {[}verb\_arg3{]}\end{tabular}} & \begin{tabular}[c]{@{}l@{}}{[}verb\_arg1{]} where column {[}verb\_arg2{]}\\  equal to {[}verb\_arg3{]} is row {[}indices of ans{]}\end{tabular} \\ 
\bottomrule[1pt]
\end{tabular}
\caption{Examples of generation templates for different operations.}
\label{verb:template}
\end{table*}

\subsection{Verbalization with Program Execution}
\label{sec:verb}

\begin{algorithm}[tb]
  \caption{Verbalization}
  \label{alg:algorithm}
  \textbf{Require}
    Statement and evidence table pair $(S, T)$, and parsed program $z^* = \{op_i\}_{i=1}^M$;
    Pre-defined operator $P=\{p_i\}_{i=1}^{R}$;
    A template function $\mathcal{F}(.)$ maps operation and operation results into sentences.
  \begin{algorithmic}[1]
    \Function {Verbalization}{$op, ret$}
    \State $args=\{\}$, $verb\_args=\{\}$ \label{line1}
    \For {$a_{j}$ in arguments of operation $op$}\label{line2}
    \If{$a_{j}$ is an operator in $P$}
    \State 
    $arg\_ans, verb\_arg$ = \Call{Verbalization}{$a_{j}, ret$} 
    \State
    $args\leftarrow args\cup arg\_ans$
    \State 
    ${verb\_args\leftarrow verb\_args\cup verb\_arg}$
    \Else
    \State 
    $args\leftarrow args\cup a_j$
    \State 
    ${verb\_args\!\leftarrow \! verb\_args \cup str(a_j)}$
    \EndIf
    \EndFor
    \State
    Apply operation $(op.{t}, args)$ over evidence table $T$, obtain operation result $ans$ \label{line13}
    \State
    Apply $\mathcal{F}(op.t, verb\_args, ans)$, obtain verbalized operation result $verb\_ans$ and verbalized operation $verb\_op$ \label{line14}
    \State
    Update $ret \leftarrow ret \cup verb\_ans$
    \State 
    \textbf{Return} $ans$, $verb\_op$
    \EndFunction
  \end{algorithmic}
  Set verbalized program execution $ret=\{\}$  \\
  \Call{Verbalization}{$op_{1}, ret$} \\
  \textbf{Return} $ret$ \\
\end{algorithm}

With the derived program $z^*$ for each $(S, T)$ pair, we propose to verbalize program execution---with operations in a program being recursively executed over the evidence table with a post-order traversal along the program tree. The verbalization algorithm works to convert the execution, including operators, arguments, and execution output, into natural language sentences, to accumulate evidences that are inherently embedded in operations. 

Formally, an operation $op_i=(op_i.t, op_i.arg)$ contains an operator $op_i.t$ and arguments $op_i.arg$, and its execution yields an output/answer $ans_i$. 
Algorithm \ref{alg:algorithm} describes the verbalization procedure. The post-order traversal over program $z^*$ and the execution of operations can be found in line~\ref{line2} to line~\ref{line13}. 
The template-based generation that converts an executed operation (its operation, the arguments, and output) into natural language sentences can be found in line~\ref{line14}. 
As such, the execution of each operation $\{op_i\}_{i=1}^M$ in the program $z^*$ is converted into an evidence sentence $V=\{v_i\}_{i=1}^M$.
Table~\ref{verb:template} lists a few operation templates.\footnote{Full templates are listed in Appendix~\ref{appendix:templates}} Note that the proposed verbalization can be easily generalized to other domains by extending templates. 
We leave it as future work for exploring different generation methods, although for structured programs with fixed operations, template-based methods are often very effective already. Fig.~\ref{fig:framework} gives an example produced by our verbalization algorithm.

\subsection{Graph-based Verification Network}
\label{sec:graph-veri}
We propose \textit{graph attention verification networks}, which is designed to fuse different sources of evidences from verbalized program execution, program structures, together with the original $S$ and table $T$, to make the final verification decision, shown on the right subfigure of Fig.~\ref{fig:framework}.

\subsubsection{Graph Definition}
\noindent

\paragraph{Nodes}
The graph $\mathcal{G}=(\mathcal{V}, \mathcal{E})$ contains three types of nodes $\mathcal{V}$ and three types of edges $\mathcal{E}$. The first type of nodes, $(n_{0},\dots,n_{M-1})$, encode verbalized program executions obtained above, called \textit{Prog-Exec} nodes, shown as green nodes on the right of Fig.~\ref{fig:framework}.
${M}$ is the number of operations in a program. The second type encodes program entities, called \textit{entity nodes}, shown as grey nodes. As each operation execution $o_i$ consists of arguments and execution output, we construct nodes $(n_{M},\dots,n_{K-2})$ for these entities. 
The third type of nodes utilize information in original tables and statements. We design a \textit{Table-BERT} node, $n_{K-1}$, initialized with the output of Table-BERT proposed in ~\cite{chen2019tabfact},
denoted as the orange node in Fig.~\ref{fig:framework}. In total, we have $K$ nodes, where $K$ varies for different ($S$, $T$, $z^*$) triples. 

\paragraph{Edges}
For a graph $\mathcal{G}$ with $K$ nodes, the adjacency matrix $A \in K \times K$ reflects their connectivity, where $A_{i,j}$ is set to 1 if node $n_i$ and $n_j$ are connected with an edge.
Similar to graph nodes, we have three types of edges. We design different attention heads to handle different types of edges/connections as detailed in Section~\ref{sec:graph_reason}. 

The first type of edges connect the nodes of verbalized program execution $V$ based on the program tree structure---we connect node $n_{i}$ and node $n_{j}$, if the corresponding operation $o_i$ is a father or child operation of operation $o_j$ in a program $z$.\footnote{We define both directed and undirected graphs; the difference is that in the directed graph, the \textit{Prog\_Exec} part of the adjacency matrix is not symmetric again. The experiments on the development set shows these two versions of graphs have similar performance, so in the remainder of the paper, we use the undirected version of the graph.} 
The second type of edges connect \textit{Prog-Exec} nodes $\{n_i\}_{i=0}^{M-1}$ with the corresponding entity nodes $(n_{M},\dots,n_{K-2})$.
The third type connects \textit{Prog-Exec} nodes, i.e., verbalized symbolic operations and executions, with \textit{Table-BERT node}, which is NLI-based verification performed directly on statement $S$ and $T$. 

\subsubsection{Graph Construction and Initialization}

For \textit{Table-BERT} node, we utilize the Table-BERT model proposed in \cite{chen2019tabfact} to obtain the representation: 
\begin{equation}
\textbf{\textit{h}}_{K-1} = f_{BERT}([\bm{\widetilde{T}};S]), \label{table-bert}
\end{equation}
where $\bm{\widetilde{T}}$ linearizes table $T$ into sentences; $\textbf{\textit{h}}_{K-1}\in\mathbb{R}^{F\times1}$ and $F$ are the number of features in node. We recommend readers to refer to~\cite{chen2019tabfact} for details. 

For \textit{Prog-Exec} nodes, instead of directly using the verbalized program executions, the nodes are constructed and initialized as follows to consider context. 
Given a program $z^*$, verbalization proposed in Sec. \ref{sec:verb} generates $M$ sentences $\{v_i\}_{i=1}^{M}$.
We use document-level BERT proposed in~\cite{liu2019text} to encode these sentences by first inserting a [\texttt{CLS}] token at the beginning of each $v_i$ and a [\texttt{SEP}] token at the end of it. The segment embedding is set to $E_A$ when $i$ is odd and $E_B$ when $i$ is even. 
After applying BERT, we take the corresponding [\texttt{CLS}] vector at the top layer (e.g., the [\texttt{CLS}] inserted before $v_2$) to be the representation for the corresponding \textit{Prog-Exec} node (e.g., $n_2$).

For \textit{entity} nodes, we take the contextualized embeddings at positions corresponding to the entities in the top layer of BERT model as the node representation. For entities with multiple words, an average pooling is applied to produce the final entity representation.  

\subsubsection{Reasoning with Graph Attentions}
\label{sec:graph_reason}
 
\paragraph{Graph Propagation} 
Unlike the standard graph propagation~\cite{velivckovic2017graph}, we model different types of edges in the propagation process. Specifically, we use the following formula to update each node representation $\textbf{\textit{h}}_i$ in graph $\mathcal{G}$:
\begin{eqnarray}
\textbf{\textit{h}}_i^{new} = f\big(\bigparallel_{d=1}^D\sigma(\sum_{j\in \mathcal{N}_i^d}\alpha_{ij}^d\bm{W}\textbf{\textit{h}}_j)\big)
\label{eq:h_i}
\end{eqnarray}
where
\begin{align}
e_{ij} &= a(\bm{U}\textbf{\textit{h}}_i, \bm{U}\textbf{\textit{h}}_j), \label{eq:e_ij}\\
\alpha_{ij}^{d} &= 
\frac{{exp}(e_{ij})}{\sum_{k=1}^K A_{i,k}^{d}{exp}(e_{ik})}
\end{align}
where $\bm{U} \in \mathbb{R}^{F\times L}, \bm{W} \in \mathbb{R}^{F\times F}$ are trainable parameters and $a(.)$ denotes shared attention mechanism $\mathbb{R}^L \times \mathbb{R}^L \rightarrow \mathbb{R}$. Note that $\bigparallel$ refers to the concatenation operation and $D$ denotes number of different types nodes ($D$ is set to 3 in this paper).  

To propagate along different types of edges, 
we extend masked attention with a multi-head mechanism to encode different types of edges in different heads. Particularly, \textit{masked attention} in self-attention mechanism is performed for each type of edges $d$. The masked attention computes normalized attention coefficients $\alpha_{i,j}^{d}$ between node $n_i$ and its neighbor $n_j$ under edge type $d$ (i.e., $A_{i,j}^{d}=1$ means node $i$ and node $j$ is connected with the edge type $d$. $A^{d}$ is the adjacency matrix we constructed above). 
To aggregate node representation from each head, we concatenate $D$ updated nodes with a feed-forward layer in Eq.~\ref{eq:h_i}, yielding the final node representation $\textbf{\textit{h}}_i^{new}$.

\paragraph{Gated Attention}
To aggregate information in a graph, we employ a gated attention mechanism to obtain final aggregated representation $\textbf{\textit{h}}_{final}$ and predict final verification label $y$ as follows:
\begin{align}
    \textbf{\textit{h}}_{final} = \sum_{i=0}^{M-1}p_i {\textbf{\textit{h}}_{i}}^{new};p_i = \sigma(\textbf{\textit{h}}_{K-1}^T {\textbf{\textit{h}}_{i}}^{new}), \label{eqn:gated_attention}\\ \centering
    y = \sigma(\bm{W_f}([\textbf{\textit{h}}_{final}\|\textbf{\textit{h}}_{K-1}]))
\end{align}
where $\bm{W_f}$ are trainable parameters, $\sigma$ is the sigmoid function, and $\|$ the concatenation operation.

\begin{table*}[ht]
\centering
\begin{tabular}{lccccc}
\toprule[1pt]
Model  & Val  & Test & Test (simple) & Test (complex) & Small Test \\ \hline
Human Performance & - & - & - & - & 92.1 \\ \hline
Table-BERT-Horizontal-S+T-Concatenate                                   & 50.7 & 50.4 & 50.8 & 50.0 & 50.3 \\
Table-BERT-Vertical-S+T-Template                                         & 56.7 & 56.2 & 59.8 & 55.0 & 56.2 \\
Table-BERT-Vertical-T+S-Template                                        & 56.7 & 57.0 & 60.6  & 54.3 & 55.5 \\
Table-BERT-Horizontal-S+T-Template                                      & 66.0 & 65.1 & 79.0  & 58.1 & 67.9 \\
Table-BERT-Horizontal-T+S-Template                                      & 66.1 & 65.1 & 79.1  & 58.2 & 68.1 \\ \hline
LPA-Voting w/o Discriminator                                            & 57.7 & 58.2 & 68.5  & 53.2 & 61.5 \\
LPA-Weighted-Voting
   & 62.5 & 63.1 & 74.6 & 57.3 & 66.8 \\
LPA-Ranking w/ Discriminator                                            & 65.2 & 65.0 & 78.4  & 58.5 & 68.6 \\
\hline
LogicalFactChecker~\cite{zhong2020logicalfactchecker} 
& 71.8 & 71.7 & 85.4 & 65.1 & 74.3
\\\hline
\textbf{ProgVGAT} & \textbf{74.9} & \textbf{74.4} & \textbf{88.3} & \textbf{67.6} & \textbf{76.2}
\\ 
\bottomrule[1pt]
\end{tabular}
\caption{Performance (accuracy) of different models on TABFACT. For \texttt{Table-BERT} baseline, different strategies of linearizing tables to bridge semantic gap with statements. \textit{Horizontal} and \textit{Vertical} refer to horizontally or vertically traverse items in tables respectively. \textit{S} denotes statements, \textit{T} denotes tables, \textit{+} indicates concatenation order between \textit{S} and \textit{T}. \textit{Concatenate} refers to directly concatenating items in tables. \textit{Template} convert items in tables into sentences with pre-defined templates. For \texttt{LPA} baseline, to select one program among all candidates for each statement, they take either a (weighted) voting strategy or a discriminator.}
\label{table:main results}
\end{table*}


\section{Experiment Setup}

\paragraph{Data and Evaluation Metric} 
As discussed above, although (semi-)structured and unstructured text data are ubiquitous in our daily life, performing fact verification across these different formats is relatively new.  We conduct our experiments on recently released large-scale dataset TABFACT~\cite{chen2019tabfact}. 

TABFACT contains 92,283, 12792, and 12779 table-statement pairs for training, validation and testing respectively. Verification on some statements requires higher-order semantics such as \textit{argmax}, the test set is further split into a \textit{simple} and \textit{complex} subset according to verification difficulty. A \textit{small} subset is provided in which the human upper-bound performance is given.
Following the existing work, we use \textit{accuracy} as the evaluation metric. Detailed data statistics are listed in Appendix~\ref{appendix:data}. 

\paragraph{Training Details}
\label{appendix:training}
For parameters in all BERT models, the hidden size is set to 768, we use Adam optimizer~\cite{kingma2014adam} with learning rate 2e-5, warmup step 3000, dropout 0.1. 
For parameters in graph attention network, the hidden feature dimensions $F$ and $L$ are all set to 768. All codes are implemented with PyTorch~\cite{NEURIPS2019_9015}. All hyper-parameters are decided according to the validation performance.

\paragraph{Compared Systems}
We compare our models with typical baselines proposed in~\cite{chen2019tabfact}. We also present results of a contemporaneous work \texttt{LogicalFactChecker}~\cite{zhong2020logicalfactchecker} which reports the best performance in the literature. Details of baselines are discussed in related work (Section \ref{relatedwork}).

\section{Results}
\paragraph{Overall Performance}
Table~\ref{table:main results} presents the results of different verification models. Our proposed method obtains accuracy of 74.4\% on the test set, achieving new state-of-the-art in this dataset.

For \texttt{Table-BERT} baseline, it leverages pre-trained language models to measure semantic similarities for table-statement pairs. 
\texttt{LPA} derives a synthesized program best describing the  statement-table pair, and executes derived program against semi-structured tables for verification. Our proposed method proposes a verbalization and graph attention network for fact verification. It integrates execution of programs into pre-trained language models, outperforming \texttt{Table-BERT} and \texttt{LPA} with a large margin.

Compared with \texttt{LogicalFactChecker}, our proposed method is built to leverage operation execution evidences and the inherent structures information with verbalization and graph attentions. 
While \texttt{LogicalFactChecker} focuses on pruning irrelevant rows and columns in evidence table with programs and utilizing structures of operations with module networks. 
Our proposed method achieves better results (74.4\%) than \texttt{LogicalFactChecker} (71.7\%). The result suggests the effectiveness of our proposed method by introducing executed operation results. Symbolic operations performed on evidence tables provide useful information for verification.

Although the proposed model improves the state-of-the-art performance on the entire dataset as well as all subsets, we can see that the \textit{complex} subset of the problem remains hard to solve. On the \textit{small} test set where the human upper-bound performance is provided (92.1\%), there is still a large gap between the system and human performance. 

\begin{table}[!htpb]
\centering
\begin{tabular}{l|cc}
\toprule[1pt]
Model               & Val  & Test  \\ \hline
Table-BERT w/ prog &70.3 & 70.0 \\ 
LogicalFactChecker  &71.8 & 71.7  \\ 
Table-BERT w/ verb. prog  &71.8 & 71.6  \\  
Table-BERT w/ verb. prog exec  & 72.4 & 72.2  \\  
\textbf{ProgVGAT} & \textbf{74.9} & \textbf{74.4}  \\ 
\bottomrule[1pt]
\end{tabular}
\caption{
Results of different ways of using operations.
}
\label{table:ablation-1}
\end{table}

\begin{table}[!htpb]
\centering
\begin{tabular}{l|ll}
\toprule[1pt]
Model    & Val  & Test  \\ \hline
\multicolumn{1}{l|}{\textbf{ProgVGAT} w/o graph attention} & 73.6 & 73.4    \\
\textbf{ProgVGAT}                & \textbf{74.9} & \textbf{74.4}  \\
\bottomrule[1pt]    
\end{tabular}
\caption{
Ablation results (accuracy) that shows the effectiveness of our graph attention component.}
\label{table:ablation-2}
\end{table}

\begin{table*}[t]
\centering
\begin{tabular}{lcc|ccc}
\toprule[1pt]
                                                                                                          & \multicolumn{1}{l}{}      & \multicolumn{1}{l|}{} & \multicolumn{3}{c}{Final Verification}                                                                          \\ \cline{4-6} 
                                                                                                          & \multicolumn{1}{l}{}      & \multicolumn{1}{l|}{} & \multicolumn{1}{c|}{Val}                   & \multicolumn{1}{c|}{Test}                  & $\Delta$Test          \\ \hline
\multicolumn{1}{l|}{\multirow{2}{*}{\begin{tabular}[c]{@{}l@{}}LPA \\ w/ CE\end{tabular}}}                & \multicolumn{1}{c|}{Val}  & Test                  & \multicolumn{1}{c|}{\multirow{2}{*}{73.3}} & \multicolumn{1}{c|}{\multirow{2}{*}{72.8}} & \multirow{2}{*}{-}    \\ \cline{2-3}
\multicolumn{1}{l|}{}                                                                                     & \multicolumn{1}{c|}{65.2} & 65.0                  & \multicolumn{1}{c|}{}                      & \multicolumn{1}{c|}{}                      &                       \\ \hline
\multicolumn{1}{l|}{\multirow{2}{*}{\begin{tabular}[c]{@{}l@{}}LPA+ BERT \\ w/ CE\end{tabular}}}          & \multicolumn{1}{c|}{Val}  & Test                  & \multicolumn{1}{c|}{\multirow{2}{*}{73.9}} & \multicolumn{1}{c|}{\multirow{2}{*}{73.4}} & \multirow{2}{*}{+0.6} \\ \cline{2-3}
\multicolumn{1}{l|}{}                                                                                     & \multicolumn{1}{c|}{67.7} & 67.3                  & \multicolumn{1}{c|}{}                      & \multicolumn{1}{c|}{}                      &                       \\ \hline
\multicolumn{1}{l|}{\multirow{2}{*}{\begin{tabular}[c]{@{}l@{}}LPA +BERT \\ w/ Margin loss\end{tabular}}} & \multicolumn{1}{c|}{Val}  & Test                  & \multicolumn{1}{c|}{\multirow{2}{*}{\textbf{74.9}}} & \multicolumn{1}{c|}{\multirow{2}{*}{\textbf{74.4}}} & \multirow{2}{*}{\textbf{+1.6}} \\ \cline{2-3}
\multicolumn{1}{l|}{}                                                                                     & \multicolumn{1}{c|}{\textbf{69.4}} & \textbf{68.5}                  & \multicolumn{1}{c|}{}                      & \multicolumn{1}{c|}{}                      &                       \\ 
\bottomrule[1pt]
\end{tabular}
\caption{Accuracy of different program selection models and corresponding final verification performance based on verbalized evidence derived from each program selection model.}
\label{table:ablation-3}
\end{table*}

\paragraph{Effect of Program Operations}
A key component of our proposed framework is utilizing the executed operation results for verification, in which we introduce a verbalization algorithm to transform the recursive execution of symbolic operations into evidence sentences. We further investigate different ways of leveraging operations: 
(1) \texttt{Table-BERT w/ prog} directly concatenates the derived program $z^*$ with the original table as new evidence, and employs \texttt{Table-BERT} on the new evidence-statement pair; (2) \texttt{Table-BERT w/ verb. prog}  differs from \texttt{Table-BERT w/ prog} in converting the derived program $z$ into sentences with templates proposed in verbalization algorithm in Sec.~\ref{sec:verb} for verification (3) \texttt{Table-BERT w/ verb. prog exec} verbalizes program along with \textit{execution} results using the algorithm proposed in Sec.~\ref{sec:verb}.

Table~\ref{table:ablation-1} shows the results. 
Compared to directly concatenate structured program with original Table-BERT, \texttt{Table-BERT w/ verb. prog} converts the structured program into natural sentences, and achieves better results. The result demonstrates the importance of eliminating the semantic discrepancy between structured data and natural language in BERT based verification model. 
\texttt{Table-BERT w/ verb. prog exec} leverages executed operation results with the verbalization algorithm and outperforms \texttt{Table-BERT w/ verb. prog} as well as \texttt{LogicalFactChecker},
The execution output provides complementary clues from evidence tables and improves the verification results. Our proposed \texttt{ProgVGAT} further leverages structures in program execution and boost the performance. The results confirm the effectiveness of leveraging symbolic operations with our method.  

\paragraph{Effect of Graph Attention}
We investigate the necessity of leveraging structure information in program execution for verification. 
We present a simpler version of our model by removing the graph attention module and integrating verbalized program execution with gated attention mechanism in Eq.~\ref{eqn:gated_attention}. Table~\ref{table:ablation-2} presents the results. By removing the graph attention network, the verification accuracy drops 1.3\% on the validation set and 1.0\% on the test set, respectively. The results show that integrating the structures of programs in the graph attention network is important for verification.

\paragraph{Effect of Derived Programs}
\label{sec:ablation-3}
We investigate the effectiveness of accurate program selection models for final verification. Table~\ref{table:ablation-3} represents the results of our model on leveraging different programs produced by different program selection models. 

We start by introducing different program selection models in Table~\ref{table:ablation-3}. \texttt{LPA w/ CE}~\cite{chen2019tabfact} applies Transformer encoders without pre-training stage (i.e., BERT) to compute semantic similarity between candidate programs and statements, and optimizes via a cross entropy loss in Eq.~\ref{program:prob}. 
It achieves a 65.0\% accuracy on the test set. Our proposed program selection denoted as \texttt{LPA+BERT w/ Margin loss}, replacing transformer encoders with BERT and optimizing the model with our proposed margin loss, can effectively improve the program accuracy to 68.5\%. 
Comparing with the accuracy of 67.3\% obtained by \texttt{LPA+BERT w/ CE}, which is optimized with cross entropy loss instead of margin loss, we can conclude that our proposed margin loss plays a positive role in program selection.

Accordingly, we compare the final verification results based on different programs selected by the above models. Our proposed method leverages programs produced by \texttt{LPA+BERT w/ Margin loss} obtains better verification results (e.g., 74.4\% on test set) compared with using programs derived by \texttt{LPA} (e.g., 72.8\% on test set). The results indicate more accurate programs can provide more useful operation information and is beneficial for the final verification. 


\begin{figure}[t]
  \centering
  \includegraphics
[width=\linewidth,trim={0cm 4.5cm 16.8cm 0cm},clip]{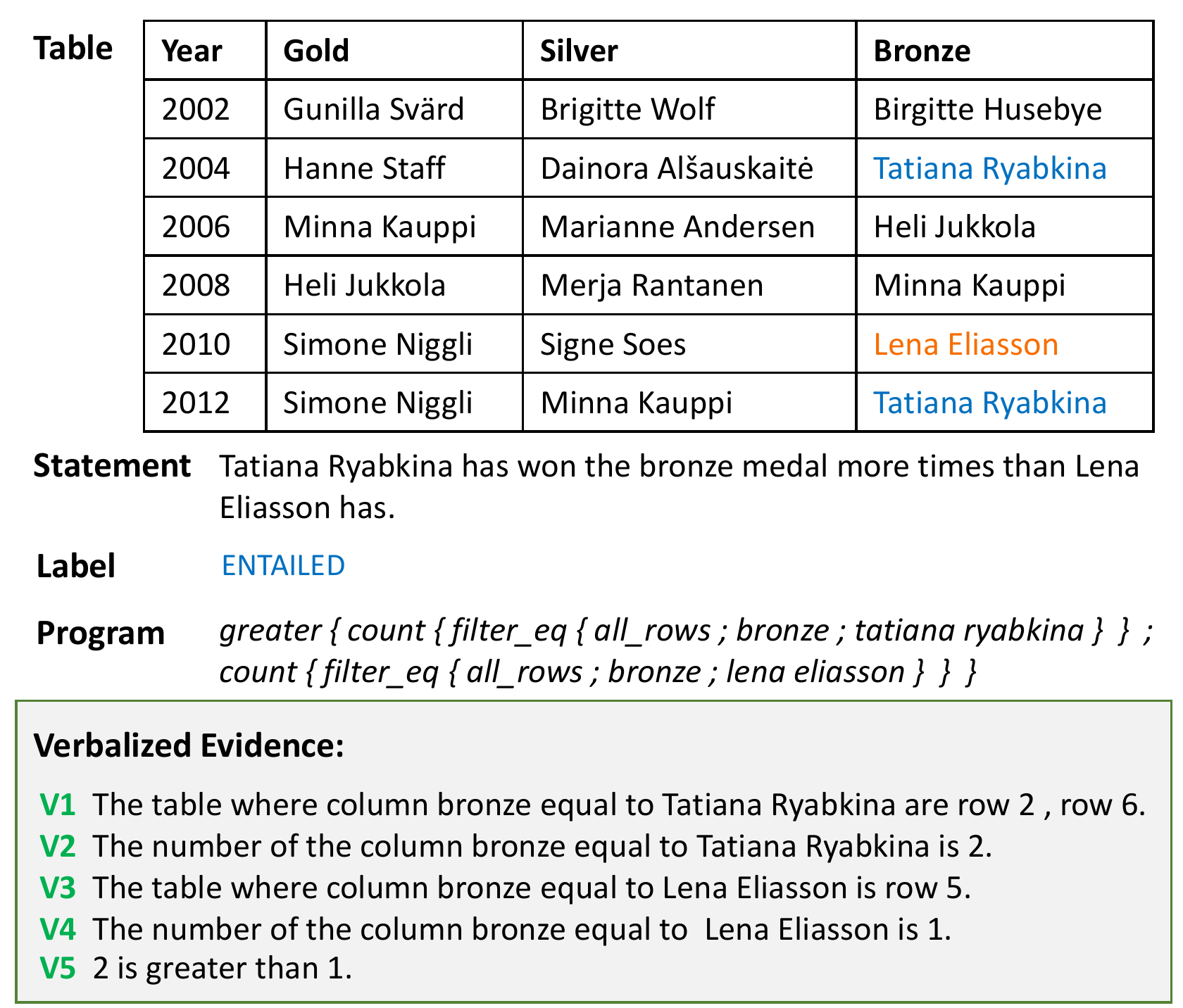}
  \caption{An example in qualitative analysis.}
\label{fig:case study}
\end{figure}

\paragraph{Qualitative Analysis} 
We provide an example, where integrating program execution information with our proposed verbalization and graph attention network can yield the right label. In Fig.~\ref{fig:case study}, the statement requires symbolic manipulation on counting bronze medals of two players and compare the number of medals of them. 
First, program selection produces a semantic-consistent program for the statement and then correctly captures the semantic correlations between the phrase ``more times than'' and operations $greater, count$. With the derived program, a verbalization algorithm executes operations over the table and produces sentences describing useful operation results. For example, ``the number of the column bronze equal to Tatiana Ryabkina is 2'', ``the number of the column bronze equal to Lena Eliasson is 1'', and ``2 is greater than 1''. Then the sentences are integrated into the verification model with the graph attention network to perform the final verification. 

\section{Conclusions}
In this paper, we propose a framework centered around \textit{programs} and \textit{execution} to provide symbolic manipulations for table fact verification. We propose a verbalization technique together with a graph-based verification network to aggregate and fuse evidences inherently embedded in programs and the original tables for fact verification. Moreover, we design a new training strategy adapting margin loss for the program selection module to derive more accurate programs. The experiments show that the proposed model improves the state-of-the-art performance to a 74.4\% accuracy on the benchmark dataset TABFACT. Our studies also reveal the importance of accurate program acquisition for improving the performance of table fact verification. 
In the future, we will investigate the properties of our proposed method on verifying statements with more complicated operations and explore the explainability of the model.

\section*{Acknowledgments}
The first, third, and last author's research is supported by NSERC Discovery Grants and Discovery Accelerator Supplement Grants (DAS). We thank Zhan Shi, Hui Liu, Jiachen Gu, and Jinpeng Wang for insightful discussions.

\bibliography{anthology,emnlp2020}
\bibliographystyle{acl_natbib}

\clearpage
\appendix

\section{Appendices}
\label{sec:appendix}
\subsection{Statistics of \textsc{TabFact} Dataset}
\label{appendix:data}
Table~\ref{table:dataset statistics} provides the statistics of \textsc{TabFact}~\cite{chen2019tabfact}, a recent large-scale table-based fact verification dataset on which we evaluate our method. Each evidence table comes along with 2 to 20 statements, and consists of 14 rows and 5-6 rows in average.

\subsection{Pre-defined Operations in Program Selection}
\label{append:operations}
Programs consists of operations, and the definition of operations are listed in Table \ref{table:operations}, mainly following \cite{chen2019tabfact}.

\subsection{Pre-defined Templates for Verbalization}
\label{appendix:templates}
In our proposed framework, there are 50 pre-defined operations, details are in Table~\ref{table:operations}. For each operation, we define templates for operation and its executed result. There are three types of executed results: (1) string or number type; (2) boolean type; (3) view or row type, where it is a sub-table or rows extracted from the evidence table. 

We represent templates for different types of operations accordingly. The detailed templates are listed in the following Table~\ref{stringtable}, Table~\ref{booltable} and Table~\ref{viewtable}. 
\begin{table}[!htpb]
\centering
\begin{tabular}{l|cccc}
\toprule[1pt]
\textbf{Split}         & \textbf{Sentence} & \textbf{Table} & \textbf{Row} & \textbf{Col} \\ \hline
Train          & 92,283     & 13,182  & 14.1 & 5.5\\
Val            & 12,792     & \phantom{0}1,696   & 14.0 & 5.4\\
Test           & 12,779     & \phantom{0}1,695   & 14.2 & 5.4\\
\bottomrule[1pt]
\end{tabular}
\caption{Statistics of \textsc{TabFact} and the split of Train/Val/Test.}
\label{table:dataset statistics}
\end{table}

\begin{table*}[!htpb]
\begin{tabular}{l|l|l|l}
\toprule[1pt]
\textbf{Operations}  & \textbf{Arguments}  & \textbf{Output}    & \textbf{Function}\\ \hline
count     & View     & Number       & Return the number of rows in the View                                                                                                                                       \\ \hline                                                                                                                                         & \begin{tabular}[c]{@{}l@{}}View, \\ Header String, \\ Cell String/\\ Number\end{tabular}     & Bool                                                     & \begin{tabular}[c]{@{}l@{}}Return whether the cell string/number\\ exists under the Header Column of the \\ given View\end{tabular}                                         \\ \hline
without                                                                                                                                        & \begin{tabular}[c]{@{}l@{}}View, \\ Header String,\\ Cell String/\\ Cell Number\end{tabular} & Bool                                                     & \begin{tabular}[c]{@{}l@{}}Return whether the cell string/number \\ does not exist under the Header Column \\ of the given view\end{tabular}                                \\ \hline
none                                                                                                                                           & String                                                                                       & Bool                                                     & \begin{tabular}[c]{@{}l@{}}Whether the string represents None, like \\ “None”, “No”, “-”, “No information provided”\end{tabular}                                            \\ \hline
before/after                                                                                                                                   & Row, Row                                                                                     & Row                                                      & Returns whether row1 is before/after row2                                                                                                                                   \\ \hline
\begin{tabular}[c]{@{}l@{}}first/second/\\ third/fourth\end{tabular}                                                                           & View, Row                                                                                    & Bool                                                     & \begin{tabular}[c]{@{}l@{}}Returns whether the row is in the first/second/\\ third position of the view\end{tabular}                                                        \\ \hline
\begin{tabular}[c]{@{}l@{}}avg/sum/\\ max/min\end{tabular}                                                                                 & \begin{tabular}[c]{@{}l@{}}View, \\ Header String\end{tabular}                               & Number                                                   & \begin{tabular}[c]{@{}l@{}}Returns the average/summation/max/min value \\ under the Header Column of the given view\end{tabular}                                            \\ \hline
\begin{tabular}[c]{@{}l@{}}argmin/\\ argmax\end{tabular}                                                                                       & \begin{tabular}[c]{@{}l@{}}View, \\ Header String\end{tabular}                               & Row                                                      & \begin{tabular}[c]{@{}l@{}}Returns the row with the max/min value under \\ the Header Column of the given view\end{tabular}                                                 \\ \hline
Hop                                                                                                                                            & \begin{tabular}[c]{@{}l@{}}Row, \\ Header String\end{tabular}                                & \begin{tabular}[c]{@{}l@{}}Number/\\ String\end{tabular} & \begin{tabular}[c]{@{}l@{}}Returns the cell value under the Header Column \\ of the given row\end{tabular}                                                                  \\ \hline
diff/add                                                                                                                                       & Number, Number                                                                               & Number                                                   & Perform arithmetic operations on two numbers                                                                                                                                \\ \hline
greater/less                                                                                                                                   & Number, Number                                                                               & Bool                                                     & \begin{tabular}[c]{@{}l@{}}Returns whether the first number is greater/less \\ than the second number\end{tabular}                                                          \\ \hline
\begin{tabular}[c]{@{}l@{}}Equal/\\ Unequal\end{tabular}                                                                                       & \begin{tabular}[c]{@{}l@{}}String, String/\\ Number, Number\end{tabular}                     & Bool                                                     & \begin{tabular}[c]{@{}l@{}}Compare two numbers or strings to see whether \\ they are the same\end{tabular}                                                                  \\ \hline
\begin{tabular}[c]{@{}l@{}}filter\_eq/\\ filter\_greater/\\ filter\_less/\\ filter\_greater\_or\_equal/\\ filter\_less\_or\_equal\end{tabular} & \begin{tabular}[c]{@{}l@{}}View, \\ Header String,\\ Number\end{tabular}                     & View                                                     & \begin{tabular}[c]{@{}l@{}}Returns the subview of the given with the cell \\ values under the Header column \\ greater/less/eq/... \\ against the given number\end{tabular} \\ \hline
\begin{tabular}[c]{@{}l@{}}all\_eq/all\_greater/\\ all\_less/\\ all\_greater\_or\_equal/\\ all\_less\_or\_equal\end{tabular}                   & \begin{tabular}[c]{@{}l@{}}View, \\ Header String,\\ Number\end{tabular}                     & Bool                                                     & \begin{tabular}[c]{@{}l@{}}Returns the whether all of the cell values under \\ the Header column are greater/less/eq/... against \\ the given number\end{tabular}           \\ \hline
and/or                                                                                                                                         & Bool, Bool                                                                                   & Bool                                                     & \begin{tabular}[c]{@{}l@{}}Returns the Boolean operation results of \\ two inputs\end{tabular}                                                                              \\ 
\bottomrule[1pt]
\end{tabular}
\caption{Details of pre-defined operations.}
\label{table:operations}
\end{table*}

\begin{table*}[!htpb]
\centering
\small
\begin{tabular}{l|ll}
\toprule[1pt]
\multicolumn{1}{c|}{\textbf{Operation}}      & \multicolumn{2}{c}{\textbf{Templates}} \\ \cline{2-3} 
\textbf{operator, {[}arguments{]}, answer} & \multicolumn{1}{c|}{\textbf{Operation}}  & \multicolumn{1}{c}{\textbf{Operation Results}}  \\ \hline
count, {[}verb\_arg1{]}, ans                                                                        & \multicolumn{1}{l|}{the number of {[}verb\_arg1{]}}                                                                                        & the number of {[}verb\_arg1{]} is {[}string(ans){]}                                                                                       \\ \hline
\begin{tabular}[c]{@{}l@{}}avg/sum/max/min,\\ {[}verb\_arg1,verb\_arg2{]}, ans\end{tabular}         & \multicolumn{1}{l|}{\begin{tabular}[c]{@{}l@{}}average/sum/maximum/minimum \\ {[}verb\_arg1{]} where column {[}verb\_arg2{]}\end{tabular}} & \begin{tabular}[c]{@{}l@{}}average/sum/maximum/minimum {[}verb\_arg1{]}\\ where column {[}verb\_arg2{]} is {[}string(ans){]}\end{tabular} \\ \hline
\begin{tabular}[c]{@{}l@{}}add/diff, {[}verb\_arg1,\\ verb\_arg2{]}, ans\end{tabular}               & \multicolumn{1}{l|}{\begin{tabular}[c]{@{}l@{}}sum/difference of {[}verb\_arg1{]}\\  and {[}verb\_arg2{]}\end{tabular}}                    & \begin{tabular}[c]{@{}l@{}}sum/difference of {[}verb\_arg1{]} and {[}verb\_arg2{]}\\ is {[}string(ans){]}\end{tabular}                    \\ \hline
\begin{tabular}[c]{@{}l@{}}uniq\_num/uniq\_string, \\ {[}verb\_arg1, verb\_arg2{]},ans\end{tabular} & \multicolumn{1}{l|}{\begin{tabular}[c]{@{}l@{}}the unique value of {[}verb\_arg1{]} \\ in column {[}verb\_arg2{]}\end{tabular}}            & \begin{tabular}[c]{@{}l@{}}the unique value of {[}verb\_arg1{]} in column \\ {[}verb\_arg2{]} is {[}string(ans){]}\end{tabular}           \\ \hline
\begin{tabular}[c]{@{}l@{}}most\_freq, \\ {[}verb\_arg1, verb\_arg2{]},ans\end{tabular}             & \multicolumn{1}{l|}{\begin{tabular}[c]{@{}l@{}}the most frequent value of {[}verb\_arg1{]} \\ in column {[}verb\_arg2{]}\end{tabular}}     & \begin{tabular}[c]{@{}l@{}}the most frequent value of {[}verb\_arg1{]} \\ in column {[}verb\_arg2{]} is {[}string(ans){]}\end{tabular}    \\ \hline
\begin{tabular}[c]{@{}l@{}}half/one\_third, \\ {[}verb\_arg1{]}, ans\end{tabular}                   & \multicolumn{1}{l|}{half/one third of value in {[}verb\_arg1{]}}                                                                           & \begin{tabular}[c]{@{}l@{}}half/one\_third of value in {[}verb\_arg1{]} is \\ {[}string(ans){]}\end{tabular}                              \\ \hline
\begin{tabular}[c]{@{}l@{}}num\_hop, \\ {[}verb\_arg1, verb\_arg2{]},ans\end{tabular}               & \multicolumn{1}{l|}{\begin{tabular}[c]{@{}l@{}}the first value of {[}verb\_arg1{]} where \\ column {[}verb\_arg2{]}\end{tabular}}          & \begin{tabular}[c]{@{}l@{}}the first value of {[}verb\_arg1{]} where \\ column {[}verb\_arg2{]} is {[}string(ans){]}\end{tabular}         \\ 
\bottomrule[1pt]
\end{tabular}
\caption{Templates for operations with string or number type executed results.}
\label{stringtable}
\end{table*}

\begin{table*}[tpb]
\centering
\small
\begin{tabular}{l|ll}
\toprule[1pt]
\multicolumn{1}{c|}{\textbf{Operation}}      & \multicolumn{2}{c}{\textbf{Templates}} \\ \cline{2-3} 
\textbf{operator, {[}arguments{]}, answer} & \multicolumn{1}{c|}{\textbf{Operation}}  & \multicolumn{1}{c}{\textbf{Operation Results}}  \\ 
\hline
\begin{tabular}[c]{@{}l@{}}only, {[}verb\_arg1{]}, \\ ans:true/false\end{tabular}                                                                      & \multicolumn{1}{l|}{number of rows in {[}verb\_arg1{]}}                                                                                                                   & \begin{tabular}[c]{@{}l@{}}number of rows in {[}verb\_arg1{]} \\ {[}true:is/false:is not{]} one\end{tabular}                                                                        \\ \hline
\begin{tabular}[c]{@{}l@{}}several, {[}verb\_arg1{]}, \\ ans:true/false\end{tabular}                                                                   & \multicolumn{1}{l|}{number of rows in {[}verb\_arg1{]}}                                                                                                                   & \begin{tabular}[c]{@{}l@{}}number of rows in {[}verb\_arg1{]} \\ {[}true:is/false:is not{]} more than one\end{tabular}                                                              \\ \hline
\begin{tabular}[c]{@{}l@{}}zero/none, {[}verb\_arg1{]}, \\ ans:true/false\end{tabular}                                                                 & \multicolumn{1}{l|}{the {[}verb\_arg1{]}}                                                                                                                                 & \begin{tabular}[c]{@{}l@{}}the {[}verb\_arg1{]} \\ {[}true:is/false:is not{]} zero/none\end{tabular}                                                                                \\ \hline
\begin{tabular}[c]{@{}l@{}}first/second/third/fourth, \\ {[}verb\_arg1, verb\_arg2{]}, \\ ans:true/false\end{tabular}                                  & \multicolumn{1}{l|}{\begin{tabular}[c]{@{}l@{}}the first/second/third/fourth row\\ {[}verb\_arg2{]} in {[}verb\_arg1{]}\end{tabular}}                                     & \begin{tabular}[c]{@{}l@{}}the first/second/third/fourth row\\ in {[}verb\_arg1{]} {[}true:is/false:is not{]} \\ row {[}verb\_arg2{]}\end{tabular}                                  \\ \hline
\begin{tabular}[c]{@{}l@{}}and/or, {[}verb\_arg1, \\ verb\_arg2{]}, ans:true/false\end{tabular}                                                        & \multicolumn{1}{l|}{{[}verb\_arg1{]} and/or {[}verb\_arg2{]}}                                                                                                             & \begin{tabular}[c]{@{}l@{}}{[}verb\_arg1{]}  and/or {[}verb\_arg2{]}\\ {[}true:is/false:is not{]} true\end{tabular}                                                                 \\ \hline
\begin{tabular}[c]{@{}l@{}}greater/less, {[}verb\_arg1,\\ verb\_arg2{]}, ans:true/false\end{tabular}                                                   & \multicolumn{1}{l|}{\begin{tabular}[c]{@{}l@{}}{[}verb\_arg1{]} greater/less\\ than {[}verb\_arg2{]}\end{tabular}}                                                        & \begin{tabular}[c]{@{}l@{}}{[}verb\_arg1{]} {[}true:is/false: is not{]}\\ greater/less than {[}verb\_arg2{]}\end{tabular}                                                           \\ \hline
\begin{tabular}[c]{@{}l@{}}equal, {[}verb\_arg1,\\ verb\_arg2{]}, ans:true/false\end{tabular}                                                          & \multicolumn{1}{l|}{\begin{tabular}[c]{@{}l@{}}{[}verb\_arg1{]} equal to\\ {[}verb\_arg2{]}\end{tabular}}                                                                 & \begin{tabular}[c]{@{}l@{}}{[}verb\_arg1{]} {[}true:is/false: is not{]}\\ equal to {[}verb\_arg2{]}\end{tabular}                                                                    \\ \hline
\begin{tabular}[c]{@{}l@{}}unequal, {[}verb\_arg1,\\ verb\_arg2{]}, ans:true/false\end{tabular}                                                        & \multicolumn{1}{l|}{\begin{tabular}[c]{@{}l@{}}{[}verb\_arg1{]} not equal to\\ {[}verb\_arg2{]}\end{tabular}}                                                             & \begin{tabular}[c]{@{}l@{}}{[}verb\_arg1{]} {[}true:is/false: is not{]}\\ not equal to {[}verb\_arg2{]}\end{tabular}                                                                \\ \hline
\begin{tabular}[c]{@{}l@{}}with/without,\\ {[}verb\_arg1, verb\_arg2, \\ verb\_arg3{]}, ans:true/false\end{tabular}                                    & \multicolumn{1}{l|}{\begin{tabular}[c]{@{}l@{}}{[}verb\_arg1{]} where column \\ {[}verb\_arg2{]} with/without value \\ {[}verb\_arg3{]}\end{tabular}}                     & \begin{tabular}[c]{@{}l@{}}{[}verb\_arg1{]} where column \\ {[}verb\_arg2{]} with/without value\\ {[}verb\_arg3{]} {[}true:is/false:is not{]} true\end{tabular}                     \\ \hline
\begin{tabular}[c]{@{}l@{}}all\_equal,\\ {[}verb\_arg1, verb\_arg2, \\ verb\_arg3{]}, ans:true/false\end{tabular}                                      & \multicolumn{1}{l|}{\begin{tabular}[c]{@{}l@{}}{[}verb\_arg1{]} where column \\ {[}verb\_arg2{]} all equal to value \\ {[}verb\_arg3{]}\end{tabular}}                     & \begin{tabular}[c]{@{}l@{}}{[}verb\_arg1{]} where column \\ {[}verb\_arg2{]} all equal to value\\ {[}verb\_arg3{]} {[}true:is/false:is not{]} true\end{tabular}                     \\ \hline
\begin{tabular}[c]{@{}l@{}}all\_less/all\_greater,\\ {[}verb\_arg1, verb\_arg2, \\ verb\_arg3{]}, ans:true/false\end{tabular}                          & \multicolumn{1}{l|}{\begin{tabular}[c]{@{}l@{}}{[}verb\_arg1{]} where column \\ {[}verb\_arg2{]} all less/greater\\ than value {[}verb\_arg3{]}\end{tabular}}             & \begin{tabular}[c]{@{}l@{}}{[}verb\_arg1{]} where column \\ {[}verb\_arg2{]} all less/greater than \\ value {[}verb\_arg3{]} {[}true:is/false:is not{]} true\end{tabular}           \\ \hline
\begin{tabular}[c]{@{}l@{}}all\_less\_or\_equal/\\ all\_greater\_or\_equal,\\ {[}verb\_arg1, verb\_arg2, \\ verb\_arg3{]}, ans:true/false\end{tabular} & \multicolumn{1}{l|}{\begin{tabular}[c]{@{}l@{}}{[}verb\_arg1{]} where column \\ {[}verb\_arg2{]} all less/greater\\ than or equal to value {[}verb\_arg3{]}\end{tabular}} & \begin{tabular}[c]{@{}l@{}}{[}verb\_arg1{]} where column {[}verb\_arg2{]}\\ all less/greater than or equal to\\ value {[}verb\_arg3{]} {[}true:is/false:is not{]} true\end{tabular} \\ 
\bottomrule[1pt]
\end{tabular}
\caption{Templates for operations with boolean type executed results.}
\label{booltable}
\end{table*}

\begin{table*}[t!]
\centering
\small
\begin{tabular}{l|ll}
\toprule[1pt] 
\multicolumn{1}{c|}{\textbf{Operation}}      & \multicolumn{2}{c}{\textbf{Templates}} \\ \cline{2-3} 
\textbf{operator, {[}arguments{]}, answer} & \multicolumn{1}{c|}{\textbf{Operation}}  & \multicolumn{1}{c}{\textbf{Operation Results}}  \\ 
\hline
\begin{tabular}[c]{@{}l@{}}before/after, \\ {[}verb\_arg1, verb\_arg2{]}, ans\end{tabular}                                                   & \multicolumn{1}{l|}{\begin{tabular}[c]{@{}l@{}}{[}verb\_arg1{]} before/after \\ {[}verb\_arg2{]}\end{tabular}}                                                        & \begin{tabular}[c]{@{}l@{}}{[}verb\_arg1{]} before/after {[}verb\_arg2{]}\\ is row {[}indices of ans{]}\end{tabular}                                                        \\ \hline
\begin{tabular}[c]{@{}l@{}}argmax/argmin,\\ {[}verb\_arg1, verb\_arg2{]}, ans\end{tabular}                                                   & \multicolumn{1}{l|}{\begin{tabular}[c]{@{}l@{}}row where column {[}verb\_arg2{]}\\ with maximum/minimum\\ value in {[}verb\_arg1{]}\end{tabular}}                     & \begin{tabular}[c]{@{}l@{}}row where column {[}verb\_arg2{]}\\ with maximum/minimum\\ value in {[}verb\_arg1{]} is row {[}indices of ans{]}\end{tabular}                    \\ \hline
\begin{tabular}[c]{@{}l@{}}filter\_eq,\\ {[}verb\_arg1, verb\_arg2, \\ verb\_arg3{]}, ans\end{tabular}                                       & \multicolumn{1}{l|}{\begin{tabular}[c]{@{}l@{}}{[}verb\_arg1{]} where column \\ {[}verb\_arg2{]} equal to value \\ {[}verb\_arg3{]}\end{tabular}}                     & \begin{tabular}[c]{@{}l@{}}{[}verb\_arg1{]} where column \\ {[}verb\_arg2{]} equal to value\\ {[}verb\_arg3{]} is row {[}indices of ans{]}\end{tabular}                     \\ \hline
\begin{tabular}[c]{@{}l@{}}filter\_less/filter\_greater,\\ {[}verb\_arg1, verb\_arg2, \\ verb\_arg3{]}, ans:true/false\end{tabular}          & \multicolumn{1}{l|}{\begin{tabular}[c]{@{}l@{}}{[}verb\_arg1{]} where column \\ {[}verb\_arg2{]} less/greater\\ than value {[}verb\_arg3{]}\end{tabular}}             & \begin{tabular}[c]{@{}l@{}}{[}verb\_arg1{]} where column \\ {[}verb\_arg2{]} less/greater than \\ value {[}verb\_arg3{]} is row {[}indices of ans{]}\end{tabular}           \\ \hline
\begin{tabular}[c]{@{}l@{}}less\_or\_equal/\\ greater\_or\_equal,\\ {[}verb\_arg1, verb\_arg2, \\ verb\_arg3{]}, ans:true/false\end{tabular} & \multicolumn{1}{l|}{\begin{tabular}[c]{@{}l@{}}{[}verb\_arg1{]} where column \\ {[}verb\_arg2{]} less/greater\\ than or equal to value {[}verb\_arg3{]}\end{tabular}} & \begin{tabular}[c]{@{}l@{}}{[}verb\_arg1{]} where column {[}verb\_arg2{]}\\ less/greater than or equal to\\ value {[}verb\_arg3{]} is row {[}indices of ans{]}\end{tabular} \\ 
\bottomrule[1pt]
\end{tabular}
\caption{Templates for operations with view or row type executed results.}
\label{viewtable}
\end{table*}

\end{document}